\documentclass{article}

\usepackage{microtype}
\usepackage{subcaption}
\usepackage{graphicx}
\usepackage{booktabs} 

\usepackage{hyperref}

 \usepackage[accepted]{icml2024}

\usepackage{amsmath}
\usepackage{amssymb}
\usepackage{mathtools}
\usepackage{amsthm}

\usepackage[capitalize,noabbrev]{cleveref}

\theoremstyle{plain}

\theoremstyle{definition}

\theoremstyle{remark}

\usepackage[textsize=tiny]{todonotes}

\title{Reasoning over the Behaviour of Objects in Video-Clips for Adverb-Type
Recognition}

\author{%
  Amrit Diggavi Seshadri \\
  Imperial College London\\
  \texttt{asd20@ic.ac.uk} \\
  \And
  Alessandra Russo\\
  Imperial College London \\
  \texttt{a.russo@imperial.ac.uk} \\
}

\begin{document}

\maketitle




\begin{abstract}
 In this work, following the intuition that adverbs describing scene-sequences are best identified by reasoning over high-level concepts of object-behaviors, we propose the design of a new framework that reasons over object-behaviour-facts extracted from raw-video-clips to recognize the clip's corresponding adverb-types. Specifically, we propose a novel pipeline that extracts object-behaviour-facts from raw video clips and propose a novel transformer based reasoning method over those extracted facts to identify adverb-types. Experiment results demonstrate that our proposed method outperforms the previous state-of-the-art on video-clips from the MSR-VTT and ActivityNet datasets. Additionally,  we release two new datasets of object-behaviour-facts extracted from raw video clips - the MSR-VTT-ASP and the ActivityNet-ASP datasets. 
\end{abstract}

\section{Introduction}
In recent years, the task of recognizing the type of actions being performed in video-clips has gained much of the vision community's attention \cite{karpathy2014large, tran2015learning,
carreira2017quo, shou2019dmc, duan2020omni, gowda2021smart, mazzia2022action, lin2023video}. It is a relatively well studied problem, with practical applications in smart-home systems and robotics. Most state-of-the-art methods for action-type recognition fuse predictions from at least two-streams of convolutional neural networks (CNNs). One stream predicts action-type probabilities from stacked image-frames of the input video-clip while another stream predicts probabilities from stacked frames of the clip's optical-flow. The output from these two streams are fused together for inference. In particular, the Inflated 3D Convolutional Network (I3D) architecture \cite{carreira2017quo} forms the backbone of many action-type recognition systems and employs this two-stream paradigm with 3D-convolutional operations.
\\\\
In contrast to action-type recognition - for which numerous architectures have been proposed, the problem of adverb-type recognition is less well explored. Adverbs further describe the nature and execution of generic action types, providing additional detail regarding intent, meaning and consequences. A device recording recipes in a kitchen for example might deem a cook in the action of ``stirring" a pot \textbf{slowly} and \textbf{completely} to be performing a required and delicate step. The same action performed \textbf{fast} or \textbf{partially} on the other hand, might be of less consequence.
\\\\
To our knowledge, there have been two architectures proposed to solve the task of adverb-type recognition in general-scene video clips \cite{doughty2020action, doughty2022you}. However, these previous methods both follow the trend set by previous action-recognition systems encoding video clips using an I3D backbone - which is not well suited for the adverbs-recognition task. 
\\\\
While end-to-end black-box CNN models such as the I3D architecture have proved successful for action-type recognition, a key reason for that success has been the fact that CNNs excel at object recognition, and a video-clip's action-type is greatly constrained by the type of objects present within scenes. The same is not true for adverb-type recognition. Adverb-types are typically \underline{not} constrained by the type of objects present within scenes and object-type may vary widely across different instances of the same adverb. A person cooking \textbf{slowly} for example presents a very different scene from a dog running \textbf{slowly} in a park. And while the use of optical-flow input does mitigate some of this problem by providing motion-related information, end-to-end CNN models such as I3D fail to generalize well over such diverse scenes. 
\\\\
However, despite this added complexity of not being constrained by object-type, humans are usually able to easily identify adverb-types by reasoning over high-level concepts of object behaviour. Something happening \textbf{slowly} for example might be identified to mean that objects change very little between frames. Properties of other adverbs such as \textbf{partially} or \textbf{completely} are harder to define, but again seem easier to identify by reasoning over higher-level concepts of object-behaviour than they are to identify by pattern-matching over diverse scenes that vary widely.
\\\\
In this work, following the intuition that adverbs are best described by reasoning over higher-level concepts, we propose a novel framework that (1)Extracts discrete object-behaviour-facts from raw video clips (2)Reasons over those extracted facts to produce high-level summaries of object-behaviour (3)Predicts and aggregates adverb-types using down-stream models over those high-level summaries.\\\\
Importantly, unlike previous work for general scene adverb-recognition \cite{doughty2020action, doughty2022you}, our framework does \underline{not} rely on I3D encodings during training or inference. Our main contributions are summarized as follows:
\begin{itemize}
    \item We propose the design of a novel framework for  adverb-type recognition in video clips that extracts object-behaviour-facts from raw video clips; reasons over those facts to learn high-level behaviour summaries; and makes predictions of adverb-types from those summaries.
    \item For the extraction phase for our framework, we propose a novel pipeline that converts raw video clips to discrete Answer Set Programs (ASP) of facts - capturing information regarding objects moving within each clip. Using this new extraction phase, we release two new datasets of object-behaviour-facts - the `MSR-VTT-ASP' and the `ActivityNet-ASP' datasets.
    \item For the reasoning phase of our framework, we propose a novel transformer-based reasoning method over our extracted ASP-facts to obtain higher-level summary vectors of object-behaviour. 
    \item Finally, we evaluate the performance of different transformer architectures within our framework against a single-step symbolic reasoning baseline and the the previous state-of-the-art.
\end{itemize}
Experiment results demonstrate that our new method for adverb-type recognition outperforms the previous state of the art on video-clips from the MSR-VTT and ActivityNet datasets, demonstrating the usefulness of a reasoning-based approach to adverb-type recognition in comparison to the previous two-stream 3D CNN based approaches. 
\section{Related Work}
\textbf{Action-Type Recognition:} Simonyan et al. \yrcite{simonyan2014two} was first to propose a two-stream 2D CNN network for action-type recognition - that employs a separate stream to process image-frames and a separate stream to process their optical flow. This two-stream method outperformed the previous method of predicting actions from features pooled across video-frame snips \cite{karpathy2014large} by a large margin. Subsequently, 3D CNNs \cite{tran2015learning} were shown to outperform their 2D counterparts by better preserving temporal information across input frame sequences, and building on these ideas, the Two-Stream Inflated 3D CNN (I3D) network was proposed \cite{carreira2017quo} - using two streams of 3D convolutional  networks over stacked frames of a video clip's image frames and optical flow. This I3D model significantly outperformed the previous methods and is employed as the backbone of a number of state-of-the-art action-type recognition systems \cite{gowda2021smart, duan2020omni, shou2019dmc}. However, as pointed out earlier, the two-stream and 3D CNN paradigms operate end-to-end, directly over raw pixel maps of image frames or optical flow and fail to cleanly separate out and reason over individual object-behaviours across time-steps.\\\\

\textbf{Adverb-Type Recognition:} Pang et al. \yrcite{pang2018human} was first to explore the problem of adverb-type recognition in video clips, introducing the ``Adverbs Describing Human Actions" (ADHA) dataset and employing a hybrid two-stream CNN along with expression detectors and human pose-estimates. However, their work addresses a problem setting different from the one that we are interested in. The ADHA dataset is focused on adverbs for human subjects, and places special focus on human pose and expression informed adverbs. We are interested in scenes comprising more general content that may not be human. Doughty et al. \yrcite{doughty2020action,doughty2022you} scaled up the problem of adverb-type recognition to general-scene video clips, and released adverb-annotations for subsets of video-clips from the HowTo100M\cite{miech2019howto100m}, VATEX\cite{wang2019vatex}, MSR-VTT\cite{xu2016msr} and ActivityNet\cite{caba2015activitynet} datasets while proposing new architectures for the task. However, as mentioned earlier, both of these prior works \cite{doughty2020action,doughty2022you} encode video-clips using an I3D backbone without attempting to reason over individual object-behaviours. Moltisanti et al. \yrcite{moltisanti2023learning} further released the Adverbs in Recipes (AIR) dataset and focused on learning adverbs in instructional recipe video-clips - which poses a different problem setting than diverse general-scenes. In our work, we use the adverb-annotations released by Doughty et al. \yrcite{doughty2022you} for experiments on general-scene video-clips from the MSR-VTT and ActivityNet datasets - where raw video-footage is available.
\section{Method}
\begin{figure}[!ht]
    \includegraphics[width=\linewidth]{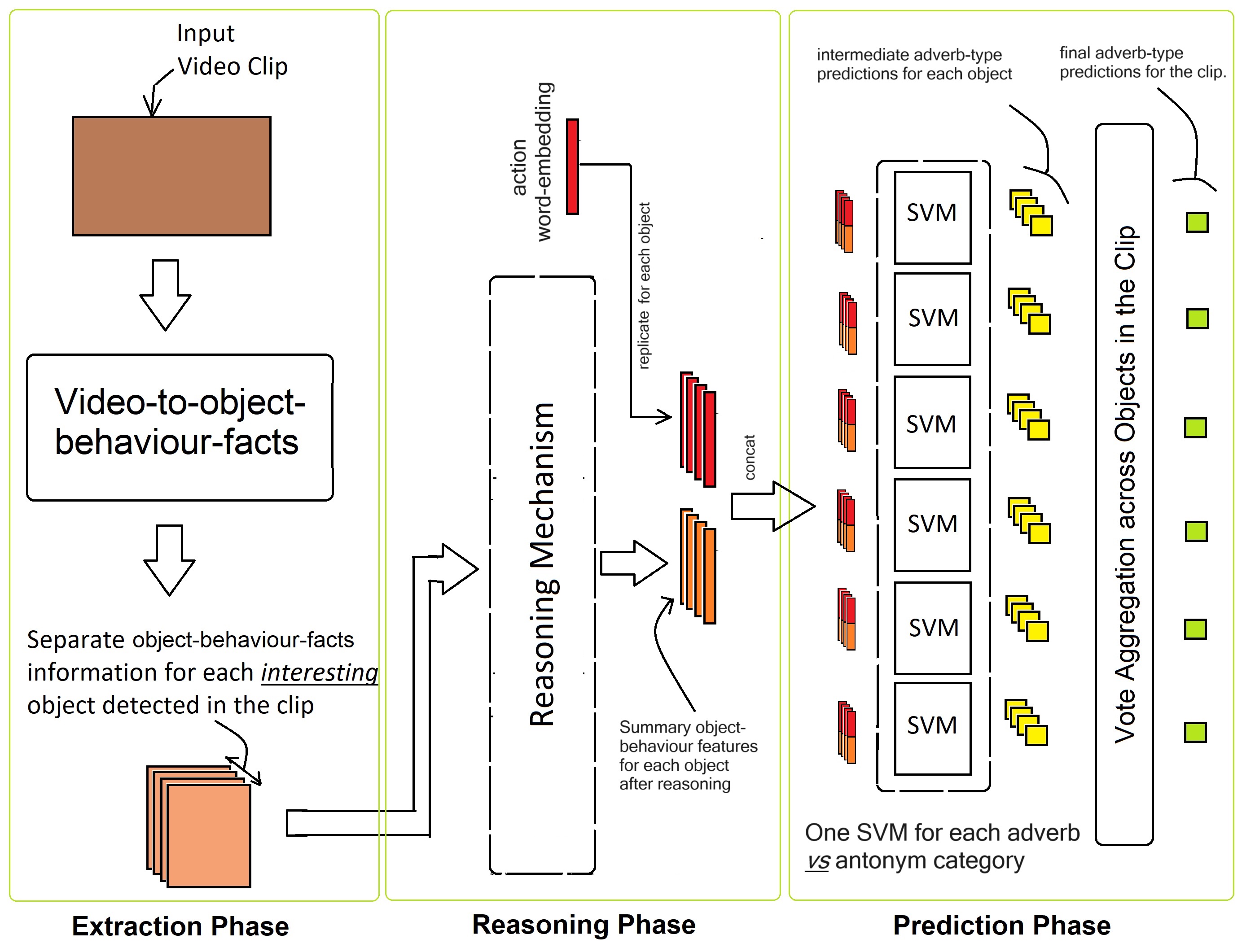}
    \caption{A depiction of our proposed framework for adverb-type recognition.}
    \label{fig:framework}
\end{figure}

Our adverb-type recognition framework (Figure \ref{fig:framework}) comprises three phases - an Extraction phase, a Reasoning phase and a Prediction phase. The Extraction phase extracts separate discrete, and human-interpretable object-behaviour-facts for each object detected to be of interest within the video clip. The Reasoning Phase summarizes those facts across time-steps into summary vectors for each object. The Prediction Phase makes downstream classifications, using separate SVMs to classify between
each adverb and it's antonym. Finally, as we obtain separate SVM predictions for each object detected to be of interest in a clip, we aggregate results by  majority-voting. 
\subsection{Extraction Phase}
\label{extraction-workflow}
\begin{figure}[!ht]
    \centering
    \includegraphics[width=\linewidth]{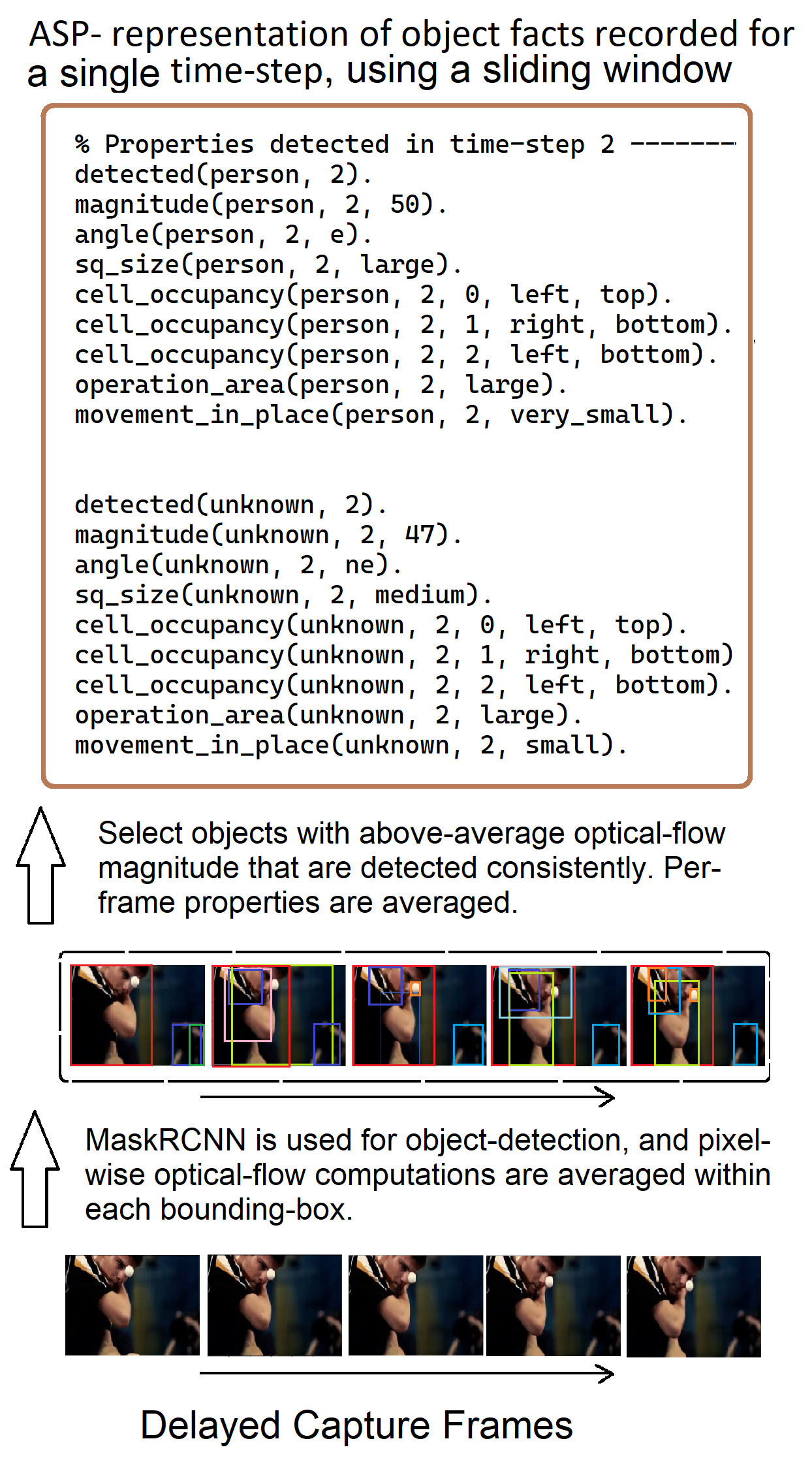}
    \caption{A depiction of our video-to-object-behaviour-facts (video-to-ASP) extraction phase.}
    \label{vid-to-ob}
\end{figure}

Figure \ref{vid-to-ob} shows a depiction of our extraction pipeline. Given a raw video clip, we first employ MaskRCNN \cite{he2017mask} over delayed-captures of static frames from the video clip's image sequence - considering every fifth frame of the original clip. Doing so, we avoid processing successive frames between which little changes. MaskRCNN gives us a collection of predicted object-types and their corresponding bounding boxes with confidence scores between 0 and 1. We ignore detections made with confidence score less than 0.3, and flag all detected patches with low confidence scores between 0.3 and 0.5 as `unknown' object-types\footnote{ To simplify our explorations and to reduce noise in the input data, in the subsequent reasoning phase, we ignore `unknown' type object-behaviours detected by our extraction phase - leaving reasoning over those less-confident object-facts as scope for future work.}. Patches detected with a confidence score above 0.5 are recorded along with their predicted object-types. 
\\
To capture properties of motion for each of these detected object patches, we compute the pixel-wise Gunnar-Farneback optical flow \cite{farneback2003two} between consecutive delayed capture frames. These per-pixel optical-flow values are averaged within each detected object-bounding-box to give us a single average numeric value of optical-flow-magnitude and a single average numeric value of optical-flow-angle for each detected object-patch. 
\\\\
To filter these numerous detections for the most adverb-relevant information, we make the assumption that faster moving objects are of more interest for adverb-type recognition then slower moving objects within the same scene.This is a reasonable assumption to make since we are trying to design a system that mimics human judgement of adverb recognition in video clips, and to humans, faster moving objects are usually more eye-catching and take precedence over slower-moving objects.\\\\
Acting on this assumption, after computing averaged optical-flow properties for each bounding box as described above, we run a non-overlapping sliding window over the delayed capture frames and remove objects that (1)Do not have optical-flow magnitude above the average of all objects detected in the same frame (2)Do not pass the first filtering step for at least half the delayed-capture frames encompassed by the sliding window. It is necessary for us to make these assumptions/choices to reduce the complexity of the problem faced by subsequent phases. Automatedly learning optimal property-extractions for adverb-type recognition is scope for future work and poses a significantly more challenging task.
\\\\
A consistently detected object of interest has its properties averaged across a time-step's window, and these properties are recorded as Answer Set Programming (ASP) \cite{lifschitz2019answer} facts as shown in Figure \ref{vid-to-ob}, where  ``detected(person, 2)" means that an object of type `person' is detected at time-step 2.
In addition to these averaged per-frame properties, we also capture local temporal properties of `operation-area' and `movement-in-place' for each object within a window. Operation-area captures the size of the area within which a single object-type `lives' for the span of a window i.e it is the product of (xmax-xmin) and (ymax-ymin) computed over all detected bounding box coordinates. Movement-in-place on the other hand is the ratio of the operation-area to average bounding-box-size of the detected object within a window. The more that an object moves around within a given window, the larger that this ratio will be. 
To simplify the down-stream reasoning process, all numeric properties besides optical-flow magnitude are placed into discrete buckets, such as `small', `very-small', `medium', etc. ; while angles are categorized into discrete sectors `north(n)', `north-east(ne)', `east(e)' and so on. The exact numerical-range of each discrete bucket/sector is specified in the accompanying code-implementation.
\begin{figure}[!h]
    \centering
    \includegraphics[width=\linewidth]{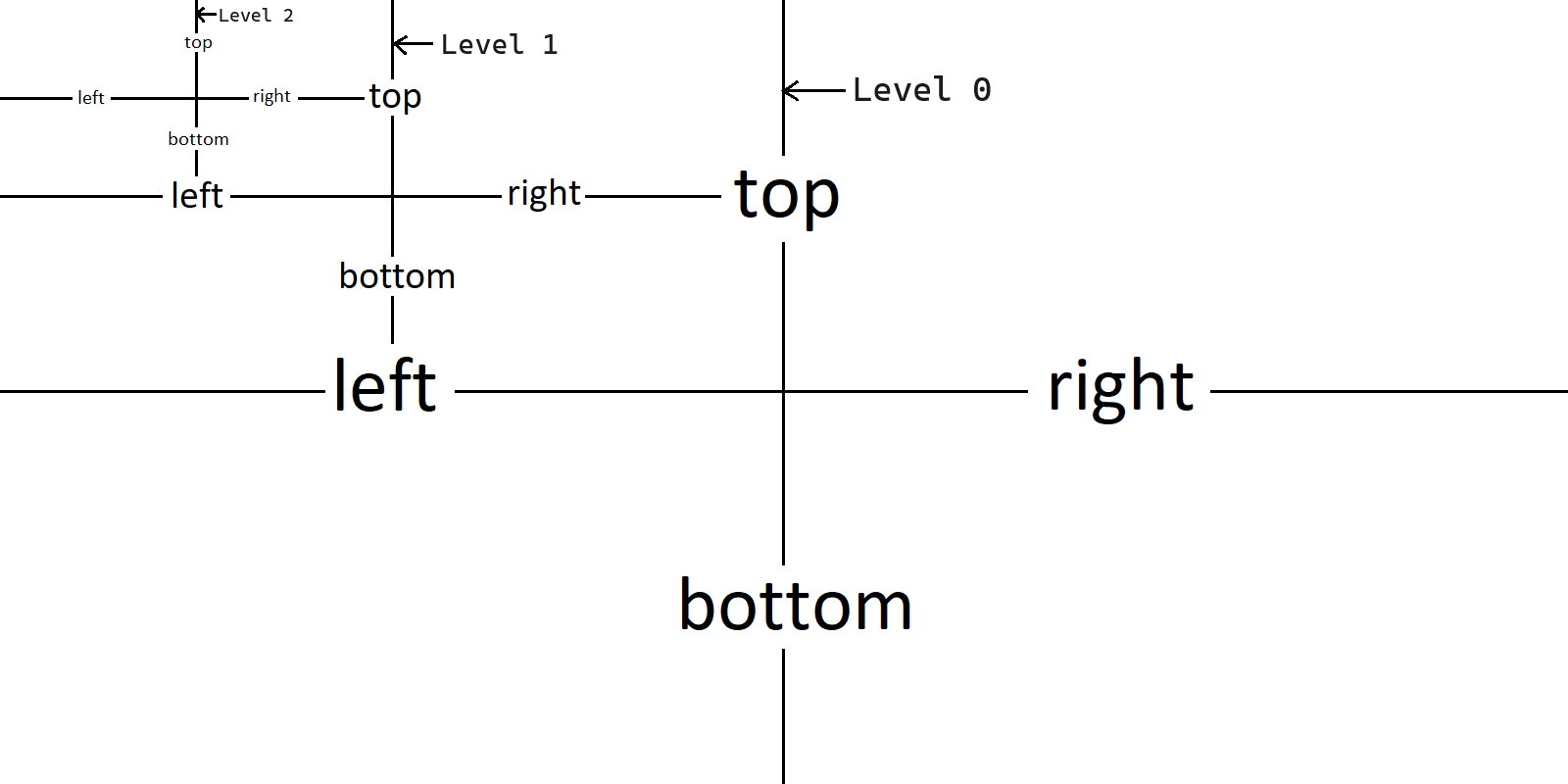}
    \caption{Hierarchy of relative-placement for cell-occupancy information.}
    \label{placement1}
\end{figure}
Besides the object properties already mentioned, we also record the region of the video frame within which each object's operation-area is maximum (i.e the section of the frame within which the object `lives'). Rather than mapping each frame's area using a series of flat grid cell locations such as $C_0,C_1,C_2,C_3...$ and assigning cell-ids to each object, in order to allow a more intuitive reasoning process, we employ a hierarchy of relative placement.
As shown in Figure \ref{placement1}, at the highest level of this hierarchy (level 0), the frame is split into 4 regions, (top, left), (top, right), (bottom, left) and (bottom, right). At the second level (level 1), each of these regions is further split into another 4 regions, and again at level 2, the process is repeated. Describing the location of an object's operation-area using this hierarchy of relative placement then allows us to make fairly simple inferences. For example, if level 1 placement stays unchanged and the level 2 toggles from left to right, we can easily determine that the object has moved right by a small amount, and if instead the level 1 placement toggles from top to bottom, then we can say that the object has moved downward by a significant amount.
\\\\
Our video-to-ASP extraction pipeline can be used to convert raw video-clips to ASP-programs in an online-fashion. However, to simplify our training and evaluation procedures and to further research efforts in symbolic and neuro-symbolic video-processing, we instead preprocess video-clips taken from the MSR-VTT and ActivityNet datasets using our proposed pipeline in an offline-manner and release new `MSR-VTT-ASP' and `ActivityNet-ASP' datasets consisting of extracted object-behaviour-facts and releavant background knowledge over predicate properties. Details of these new datasets are discussed further in Sections \ref{datasets}. 
\subsection{Reasoning Phase}
\subsubsection{Single-Step Symbolic-Based Baseline}
\label{sym-reasoning}
\begin{figure}
\centering
    \begin{subfigure}[b]{0.9\linewidth}
    \centering
    \includegraphics[width=\linewidth]{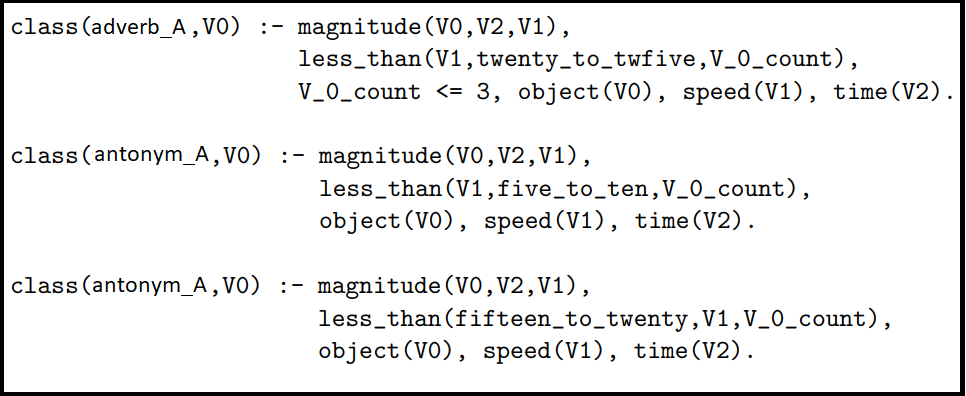}
    \caption{Batch-plausible ASP range-rules learnt to logically-explain some (adverb\_A, antonym\_A) categories.}
    \label{toy-eg-rules}
    \end{subfigure}
    \begin{subfigure}[b]{0.9\linewidth}
    \centering
    \includegraphics[width=\linewidth]{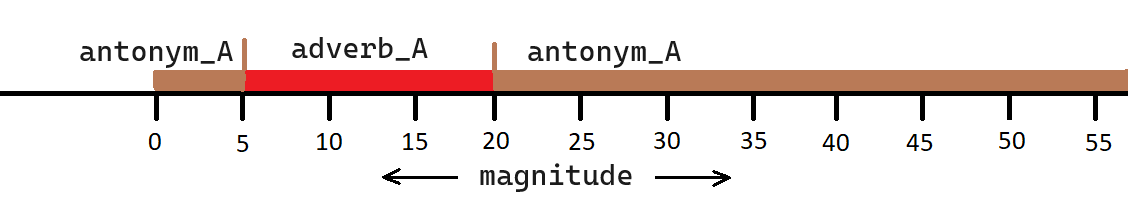}
    \caption{A depiction of the ranges defined by the rules shown in Figure \ref{toy-eg-rules}.}
    \label{ranges}
    \end{subfigure}
    \caption{A depiction of rules learnt by FastLAS for some batch of example object-behaviours labeled with (adverb\_A, antonym\_A) categories.}
\end{figure}

In this work, we first consider employing the FastLAS inductive learning method \cite{law2020fastlas} to automatedly learn indicator rules that plausibly define adverb-types. However,  automatedly learning symbolic rules that reason over more than one time-step is an extremely challenging task (owing to the large number of possible variable groundings of rules). Rather than attempting to overcome the multi-time-step symbolic challenge, in this initial exploration, we use single-step symbolic-reasoning as a baseline approach and consider learning a large number of simple single-time-step rules that compositionally might inform overall adverb-type.
In particular, for \textit{magnitude}, \textit{angle}, and \textit{operation\_area} predicates, we focus on learning range-rules that define upper or lower bounds of an object's predicate-properties at single time-steps, and for the \textit{cell-occupancy} predicate, we focus on learning single-time step rules that outline rough left/right or up/down placement of an object within the frame.
As an example, Figures \ref{toy-eg-rules} and \ref{ranges} depict FastLAS learnt ASP range-rules that classify between some categories of `adverb\_A' and `antonym\_A' motion for some collection of object-behaviour facts. According to these indicator rules, objects moving with an optical-flow magnitude between five and twenty at some time-step are considered to exhibit `adverb\_A' behavior, while those objects moving with optical-flow magnitude outside those limits are considered to exhibit `antonym\_A' behavior. These rules might not hold as universally true, however they are identified by FastLAS as plausibly explanations for some given batch of input behaviour-examples.
Similarly, from the training data, we learn indicator rules classifying between each (adverb, antonym) pair. To do so we sample small-balanced-batches of object-behaviour-facts from the training data - choosing 10 randomly sampled object-behaviours for each adverb, and 10 random sampled object-behaviours for it's antonym. We then run FastLAS separately over each balanced-batch along with common background-knowledge to obtain a large number of batch-wise plausible adverb/antonym indicator-rules over predicate-properties (such as those rules shown in Figure \ref{toy-eg-rules}).
After all such single-time-step batch-plausible indicator-rules have been learnt for each adverb vs antonym task across the training data, we use those symbolic ASP rules to summarize object behaviours. Specifically, for an object's collection of behaviour-facts, we assign a 1 for an indicator-rule if that rule logically-fires for the given object's behaviour-facts, and we assign a 0 otherwise - so that from our collection of indicator-rules we obtain a vector of 0s and 1s (such as [1,1,1,0,1,1,...]) for each object-behaviour. All object-behaviours are converted in this manner for each adverb vs antonym task, and those vectors are used as rough behaviour-summaries for downstream adverb-type recognition. (Implementation details are further discussed in Appendix Section A.2).

\subsubsection{Transformer-Based Reasonoing}
\label{transformer-reasoning}
\begin{figure}[htp]
    \centering
    \begin{subfigure}[b]{0.9\linewidth}
    \centering
    \includegraphics[width=\linewidth]{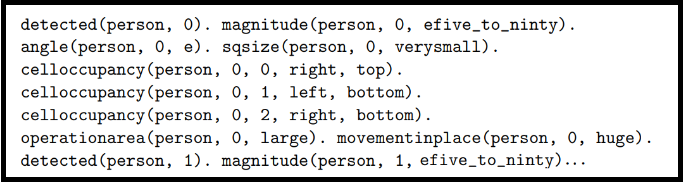}
    \caption{A truncated object-behaviour snippet showing ASP facts of object-behaviour for a `person' type object across two time steps - formatted for symbolic-based reasoning.}
    \label{snippet}
    \end{subfigure}
    \begin{subfigure}[b]{0.9\linewidth}
    \centering
    \includegraphics[width=\linewidth]{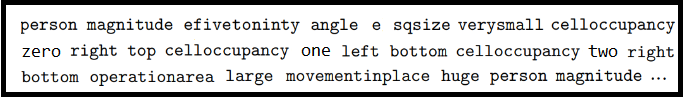}
    \caption{A truncated flattened representation of the object behaviour from Figure \ref{snippet} - formatted for transformer-based reasoning.}
    \label{flat}
    \end{subfigure}
    \caption{Object Behaviour for a `person' type object formatted for symbolic (a) and transformer (b) based reasoning.}
    \label{behaviours}
\end{figure}

\begin{figure}[htp]
    \centering
    \includegraphics[width=0.8\linewidth]{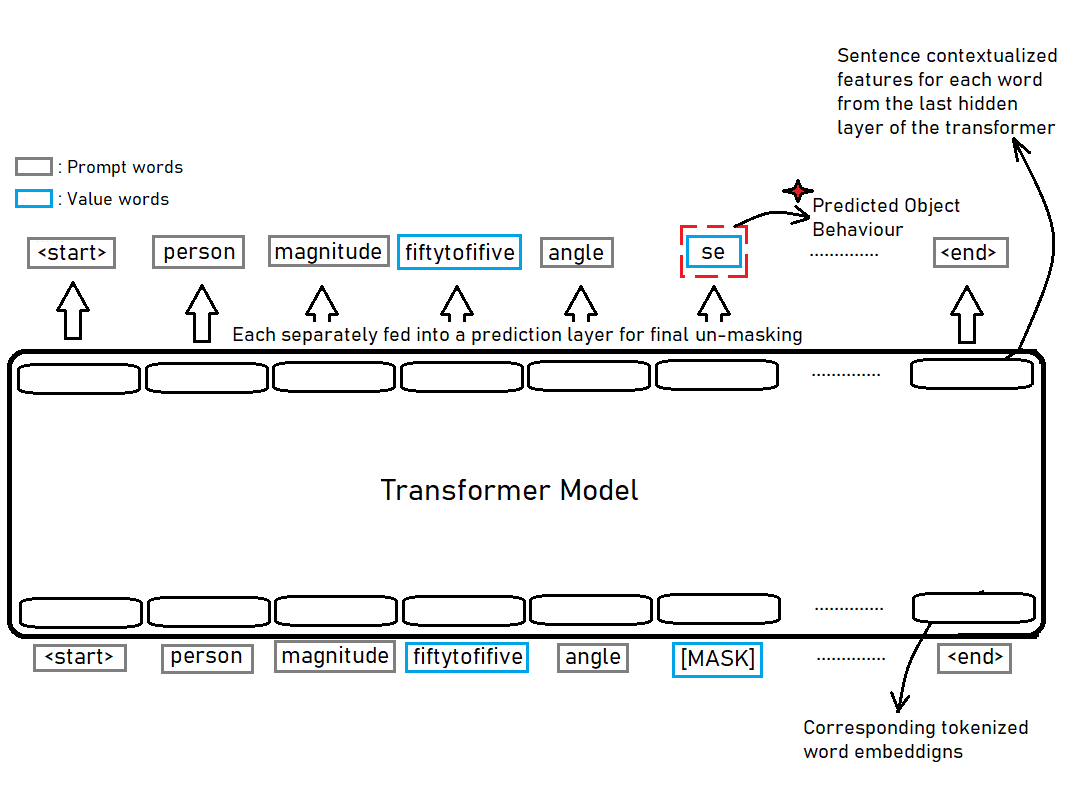}
    \caption{Custom Masked Language Modeling over value-words of object-behaviours.}
    \label{mlm-mask}
\end{figure}

Alternative to our single-time-step symbolic-based baseline, in this work, we propose a novel multi-time-step transformer-based approach to reason over object-behaviors. We start by flattening the ASP-format object-behaviour properties detected by our Extraction Phase (as shown in Figure \ref{behaviours}). We get rid of unnecessary syntactical detail and special characters that might otherwise confuse a sentence tokenizer, and record object-type only once per time-step to avoid redundancy. We also eliminate the explicit time-stamps $(1,2,3...)$ associated with each logical fact. We are able to do this, provided that we maintain the correct chronological ordering of detected object-properties since transformer models already have provisions allowing them to recognize and reason over the positional-ordering of words in sentences.
Next, we consider a custom Masked Language Modeling (MLM) \cite{devlin2018bert} approach over object-behaviours to learn useful object-behaviour representations (Figure \ref{mlm-mask}). In conventional MLM, some of the words of a natural-language sentence are masked out and a transformer model fitted with a shallow prediction-layer is trained to predict those masked words from the rest of the unmasked sentence - forcing the transformer to learn to encode sentence-structure and overall sentence meaning. Features output by the last transformer-layer are then typically extracted and used for related down-stream tasks such as text-classification. In the context of our object-behaviours problem setting, we adapt this idea, by masking out some of the `value-words' that correspond to each object's particular behaviour (that might be a value of magnitude/angle/operation-area/etc. at some time-steps). We then train a transformer model to predict those masked values from the rest of the unmasked object-behaviour (as shown in Figure \ref{mlm-mask}) \footnote{Specifically, we mask value-words with a probability of $20\%$ and do \underline{not} mask prompt-words such as `magnitude' and `angle' that occur in every example. Importantly, we also make sure \underline{not} to mask object-types as they can be difficult to infer from object-motion, and forcing a model to predict them would detract from learning other behavioural-properties.}. In doing so, we force the transformer to learn to encode some overall meaning or dynamics of object-behaviour. The features output by the last transformer-layer are then used for down-stream adverb-type recognition. 
In particular, we do not train the transformer model from scratch, but rather fine-tune a model that has been pretrained for natural-language MLM. We do this transfer-learning in order to exploit complex network reasoning properties that have already been learnt over very large datasets of natural language\footnote{Note: to limit the computational costs of fine-tuning, we truncate flat object behaviour inputs (Figure \ref{flat}) at 512 tokens.}.
Once we have fine-tuned our MLM model to reason over and unmask object-behaviours, we then use it to extract object-behaviour summary vectors. For each input object-behavior snippet in the dataset, we feed that flattened object-behavior to our trained transformer-model and extract the word-level vectors output by the transformer's final layer. Those word-level features are then averaged across the entire flattened sentence to give us a single summary vector - that encodes some overall multi-time-step object-behaviour information.
\subsection{Prediction Phase}
As in previous work \cite{doughty2020action, doughty2022you}, our adverb-type prediction is  conditioned on the video-clip's overall action-type. Each summary object-behaviour vector (output by an object-behaviour reasoning approach) is concatenated with a word2vec \cite{mikolov2013efficient} embedding of the video-clip's action-type and fed into a separate Support-Vector Machine (SVM) with rbf kernel for binary classification between each adverb-type and it's antonym. At test time, the adverb-vs-antonym predictions from multiple object-behaviours detected to be of interest in a single clip are aggregated by a majority-vote. 
\section{Experiments}
\label{datasets}
\textbf{Datasets:} We evaluate our method on subsets of the MSR-VTT \cite{xu2016msr} and ActivityNet \cite{caba2015activitynet} datasets, using adverb-annotations by Doughty et al. \cite{doughty2022you}. We process clips where both raw-footage and adverb-annotations are available using our Extraction Phase (Section \ref{extraction-workflow}), to obtain 1309 ASP-programs for our new MSR-VTT-ASP dataset and 1751 ASP-programs for our new ActivityNet-ASP dataset - where each program contains facts of multiple object-behaviours detected to be of interest within the corresponding video-clip, along with background knowledge of predicate properties (Appendix 6.1). Each program is labeled with one or more of 22 adverb-types (11 adverb/antonym pairs) according the source clip's labels\footnote{We drop the loudly/quietly category since neither our method nor the previous work uses a clip's audio.} : (1)upwards/downwards, (2)forwards/backwards, (3) outdoor/indoor, (4) slowly/quickly, (5)gently/firmly, (6)out/in, (7) partially/completely, (8)properly/improperly, (9) periodically/continuously, (10) instantly/gradually, (11) off/on. This leaves us with 1674 unique (asp-program, adverb) pairs from MSR-VTT and 1824 unique (asp-program, adverb) pairs from ActvityNet. We randomly split these datasets into training and testing sets using $70/30$ stratified splits (stratified by adverb-type) to obtain 1171 training and 503 testing samples for MSR-VTT-ASP and 1276 training and 548 testing samples for ActivityNet-ASP. Summary statistics for these two new datasets are shown in Table \ref{datasets-summary-table1}.
\\\\
Finally, with these two new ASP-datasets and splits having been created, for experiments, we turn to the requirements of our adverb-type recognition framework. We require snippets of individual object-behaviours to reason-over for adverb-type prediction. So, for each ASP-program, we cut-out behaviour snippets for separate detected objects - so that one snippet corresponds to one object's behaviour over the course of a video (as shown in Figure \ref{snippet}). Each object behaviour snippet is annotated with the adverb-type of it's source program. These snippets of object-behaviour are then repeated within each adverb-category to balance out the number of samples used for training and testing in each category.
\begin{table*}[!ht]
\centering
\begin{tabular}{llllll}
                &&&&\hspace*{1cm} Average.\\
                \cmidrule(r){4-6}
                & \# Train Clips & \# Test Clips & \# objects per clip & \# time-steps per clip & \# objects per time-step \\
            \midrule
MSR-VTT-ASP     & 1171          & 503          &  3.06                           &   16.35                     &      0.93                          \\
ActivityNet-ASP & 1276          & 548          &  3.24                           &   29.61                     &  0.87    
\end{tabular}
\caption{Summary of properties of our new MSR-VTT-ASP and ActivityNet-ASP datasets of object-behaviour-facts - Ignoring `unknown' types and assuming that an object-type corresponds to the same physical-object across all detections within a clip.}
\label{datasets-summary-table1}
\end{table*}
\subsection{Single-Step Symbolic Based Baseline} 
As mentioned in section Section \ref{sym-reasoning}, we learn a large number of batch-wise plausible indicator rules using FastLAS over balanced batches of object-behaviour-facts from the training-set within each adverb-vs-antonym category, and use those learnt indicator-rules to extract summary vectors of object behaviors for each adverb-vs-antonym classification task. Those summary vectors from the training-set are then concatenated with word2vec embeddings of the video-clip's action type and used to train separate SVM classifiers to distinguish between each adverb and its antonym. At test time, SVM predictions from multiple object-behaviours within individual source-video-clips from the test-set are aggregated by majority-vote to distinguish between adverbs and antonyms in each category. As shown in Figure \ref{bar-chart}, the accuracy of prediction of this single-time-step based symbolic method is highest for both MSR-VTT-ASP and ActivityNet-ASP datasets when distinguishing between gently-and-firmly type adverbs - suggesting that this category might plausibly be inferred from a grouping of single-time-step behaviour-properties. Performance is worst for more complex adverb-types: periodically-continuously, instantly-gradually and off-on - for which no single-time-step batch-wise-plausible range rules are found and adverb-type predictions are made only from the action-embedding. Rules learnt are further discussed in Appendix Section A.3. 
\subsection{Transformer Based Reasoning}
\begin{figure*}[!htp]
    \centering
    \begin{subfigure}[b]{0.45\textwidth}
    \centering
    \includegraphics[width=\linewidth]{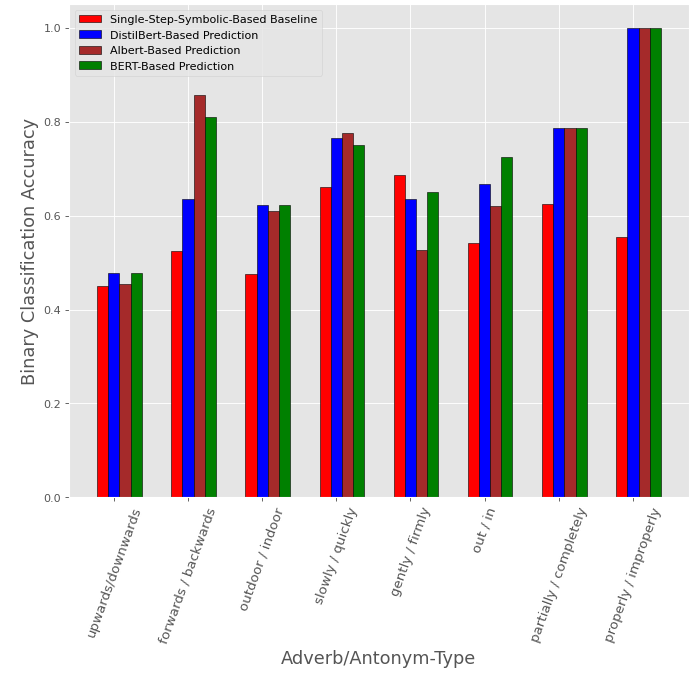}
    \caption{MSR-VTT-ASP}
    \end{subfigure}
     \begin{subfigure}[b]{0.50\textwidth}
    \centering
    \includegraphics[width=\linewidth]{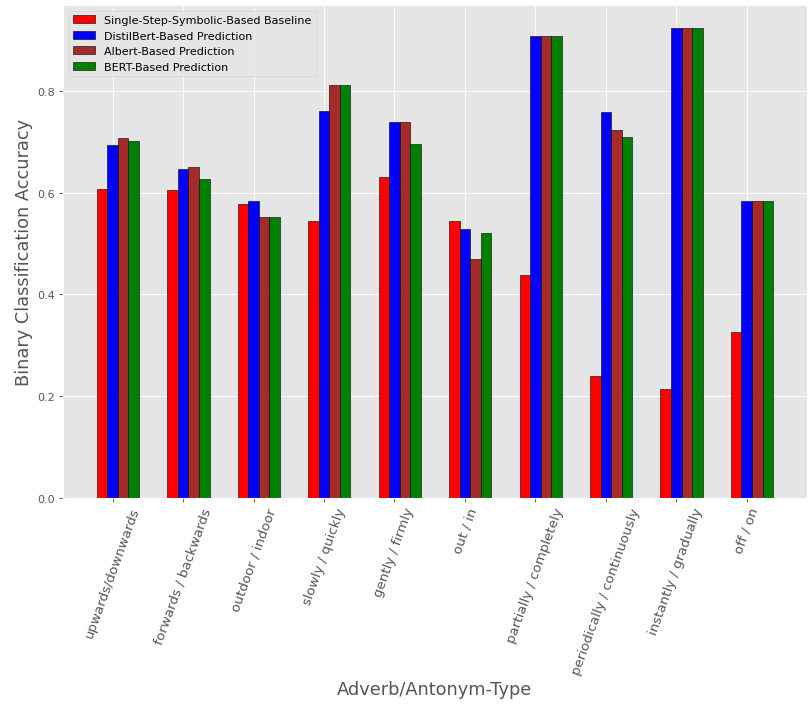}
    \caption{ActivityNet-ASP}
    \end{subfigure}
    \caption{Binary classification accuracies for adverb-vs-antonym tasks on the ASP test-sets of balanced object-behaviours using our proposed symbolic and transformer-based approaches.}
    \label{bar-chart}
\end{figure*}

For our experiments, we consider three variants of the landmark BERT \cite{devlin2018bert} transformer architecure - namely the full large BERT model \cite{devlin2018bert}, and the smaller ALBERT  \cite{lan2019albert} and DistilBERT \cite{sanh2019distilbert} models. As outlined in Section \ref{transformer-reasoning}, we fine-tune each pre-trained transformer model by flattening  object-behaviour snippets and making them unmask randomly masked `value-words'. We then obtain behaviour summary-vectors for each object-snippet by feeding their flat representations to the trained transformers without masking and average word-level features output by the last hidden layer across the entire sentence. As with the symbolic-case, we concatenate action-type word-embeddings and a separate SVM with rbf kernel is used over these vectors, along with majority-voting to distinguish between each adverb and it's antonym. We find that DistilBERT, ALBERT and BERT achieve strong average-performances, while out-performing our symbolic baseline by a wide margin (Table \ref{avg-accuracy}). When each adverb-vs-antonym recognition task is viewed separately (Figure \ref{bar-chart}), we note that transformer-reasoning performance is tied for the three transformer-models across a number of the tasks. The symbolic baseline performs best in the `gently/firmly' category in MSR-VTT-ASP and 'out/in' category in ActivityNet-ASP, while one or the other of our transformer-based methods works best for other adverbs. The general superiority of the transformer approach is largely to be expected, given that it jointly reasons over multiple time-steps and multiple predicate properties, while our single step symbolic baseline composes single time-step, single predicate properties. It can be difficult to interpret why one transformer reasoning method outperforms another within a given category. However, it is encouraging to note that not all models exhibit the same performance, since we can achieve higher overall accuracy for adverb-type recognition by altering transformer model architetures.  
\subsection{Comparison with State-of-the-Art}
We retrain the previous I3D dependent methods on the clips corresponding to our MSR-VTT-ASP and ActivityNet-ASP dataset splits to make a comparison against the previous state-of-the-art \cite{doughty2020action,doughty2022you}. In the case of both datasets, all three of our transformer-reasoning approaches are highly competitive and outperform the previous state-of-the-art, with BERT-based reasoning achieving the highest average accuracy for MSR-VTT-ASP - marking a 5.26\% improvement over PsudoAdverbs \cite{doughty2022you}, and DistilBERT-based reasoning achieving the highest average accuracy for ActivityNet-ASP - marking a 3.67\% improvement over PsudoAdverbs, demonstrating the usefulness of a reasoning-based approach over object-behaviours for adverb-type recognition in comparison to the previous two-stream 3D CNN approaches.  
\begin{table*}[!h]
    \centering
    \begin{tabular}{lll} 
        \toprule
        & MSR-VTT-ASP & ActivityNet-ASP  \\
        Method  & Average accuracy (\%) $\uparrow$ & Average accuracy (\%) $\uparrow$ \\
        \cmidrule(r){1-3} 
        ActionModifiers \cite{doughty2020action}& 59.98& 61.05 \\
        PsudoAdverbs \cite{doughty2022you} & 67.49 & 67.59\\
        \cmidrule(r){1-3} 
        \cmidrule(r){1-3} 
        \textbf{Ours:} & &\\
        \cmidrule(r){1-3} 
         Single-Step Symbolic-Based Baseline &  56.49& 47.25\\
         \cmidrule(r){1-3} 
         DistilBERT Transformer-Based Prediction & 69.88 & \textbf{71.26}\\
         ALBERT Transformer-Based Prediction & 70.37 & 70.73 \\
         BERT Transformer-Based Prediction & \textbf{72.75} & 70.35 \\
     \bottomrule
    \end{tabular}
    \caption{Average adverb-vs-antonym binary-classification accuracies using our proposed reasoning methods on the balanced ASP test-sets, and the previous state-of-the-art on corresponding clips.}
    \label{avg-accuracy}
\end{table*}\\\\
\textbf{Broader Impact:} Similar to action-type recognition, we reflect that one may attempt to use adverb-type recognition to maliciously monitor video-footage. However, we also note that improved adverb-type recognition, when used ethically for better video-interpretation, offers significant benefits to human computer interaction and robotics.  \\\\
\textbf{Limitations and Scope for Future Work: } To our knowledge, we are the first to reason-over object-behaviours for adverb-type recognition. As such there are several possible directions for further investigation. Primarily, our extraction-phase was not optimally learnt from data, for transformers we explored custom MLM-modeling using BERT-style transformer models, and our symbolic-baseline was limited to single-time step and single-predicate type rules. Scope for future work then includes automatedly learning optimal object-behaviour-fact extractions, exploring alternative transformer-modeling/architectures (such as causal modeling using GPT-3 \cite{brown2020language}), and reasoning over multiple time-steps and multiple predicate-properties using neuro-symbolic reasoning. 
\section{Conclusion}
 In this work, we proposed the design of a new framework that reasons over object-behaviour-facts extracted from raw-video-clips to recognize the clip's corresponding adverb-types. We proposed a novel pipeline to extract object-behaviour-facts from raw video clips and released two new datasets of object-behaviour-facts - the ‘MSR-VTT-ASP’ and the ‘ActivityNet-ASP’ datasets. We proposed a novel transformer-based reasoning method over those extracted facts to identify adverb-types. Experiment results demonstrated that our new method outperforms the previous state-of-the-art on video-clips from the MSR-VTT and ActivityNet datasets - demonstrating the usefulness of a reasoning-based approach to adverb-type recognition.

\bibliography{main}
\bibliographystyle{icml2024}

\clearpage
\appendix
\section{Appendix}

\subsection{Background Knowledge of Predicate properties}
\label{extraction-appendix}
\begin{figure}[!h]
    \centering
    \includegraphics[width=\linewidth]{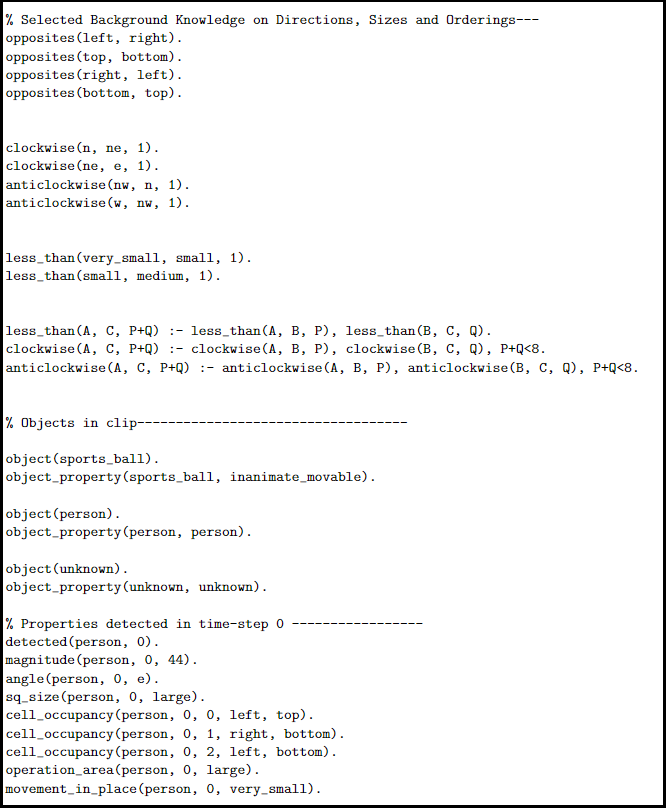}
    \caption{A trimmed example of object-behaviour captured in an ASP program with selected background knowledge.}
    \label{trim_clip_prog}
\end{figure}

For clarity, Figure \ref{trim_clip_prog} shows a trimmed example of an ASP-program, computed from a video-clip using our extraction pipeline (Section \ref{extraction-workflow}). The figure shows some selected background knowledge and object-properties detected over a single time step. Importantly, we highlight that our background knowledge includes information on the `opposites' of relative directions, as well as `less-than', `clockwise' and `anticlockwise' orderings over bucketed numeric values - so that we can reason about ranges in symbolic approaches. Importantly, these ordering predicates are formulated with a measure of distance to allow for range-reasoning. For example, `clockwise(n, ne, 1)' indicates that northeast is clockwise of north by one-tick while `clockwise(n, e, 2)' indicates that east is clockwise from north by two-ticks. Similarly `less-than(very-small, small, 1)' indicates that small is larger than very-small by one step, whereas `less-than(very-small, medium, 2)' indicates that medium is larger than very-small by two steps.
\\
We also include transitive rules, logically defining that if A is \textit{$less\_than$} B by P steps, and B is \textit{$less\_than$} C by Q steps, then A is \textit{$less\_than$} C by P+Q steps.  \\\\
Similarly, if B is \textit{$clockwise$} of A by P ticks, and C is \textit{$clockwise$} of B by Q ticks, then C is \textit{$clockwise$} of A by P+Q ticks, and the same for the $anticlockwise$ property. (Note: We limit the transitive $clockwise$ / $anticlockwise$ properties to a range of 8 ticks to avoid learning cyclic rules).

\subsection{Implementation Details of the Single-Step Symbolic-Based Baseline}
\begin{figure}[htp]
    \centering
    \includegraphics[width=\linewidth]{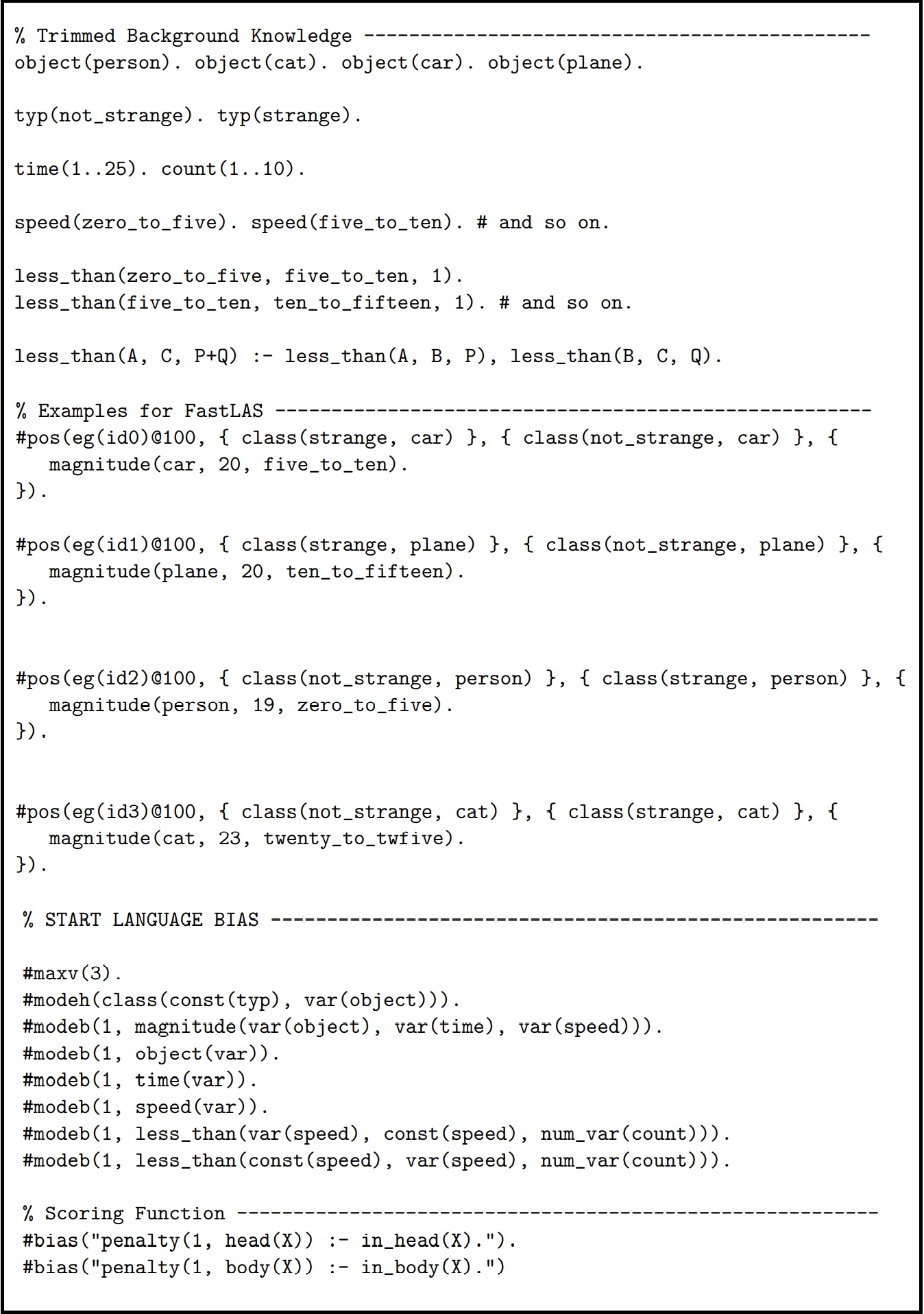}
    \caption{Toy example of range-based rule-learning.}
    \label{toy-eg}
\end{figure}

\begin{figure}[htp]
    \centering
    \includegraphics[width=\linewidth]{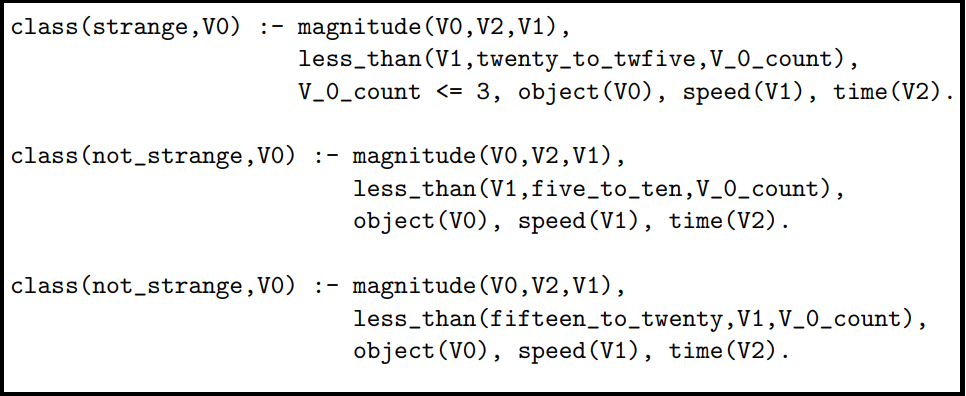}
    \caption{Rules learnt by FastLAS for the toy-example shown earlier.}
    \label{toy-eg-rulesA}
\end{figure}

\begin{figure}[htp]
    \centering
    \includegraphics[width=\linewidth]{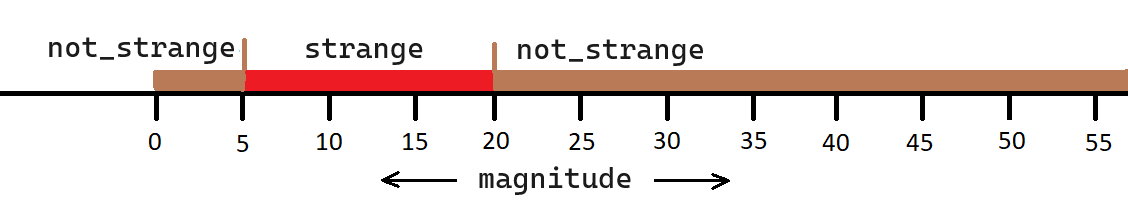}
    \caption{A depiction of the ranges defined by the rules learnt for our toy-example.}
    \label{rangesA}
\end{figure}

\begin{figure}[htp]
    \centering
    \includegraphics[width=\linewidth]{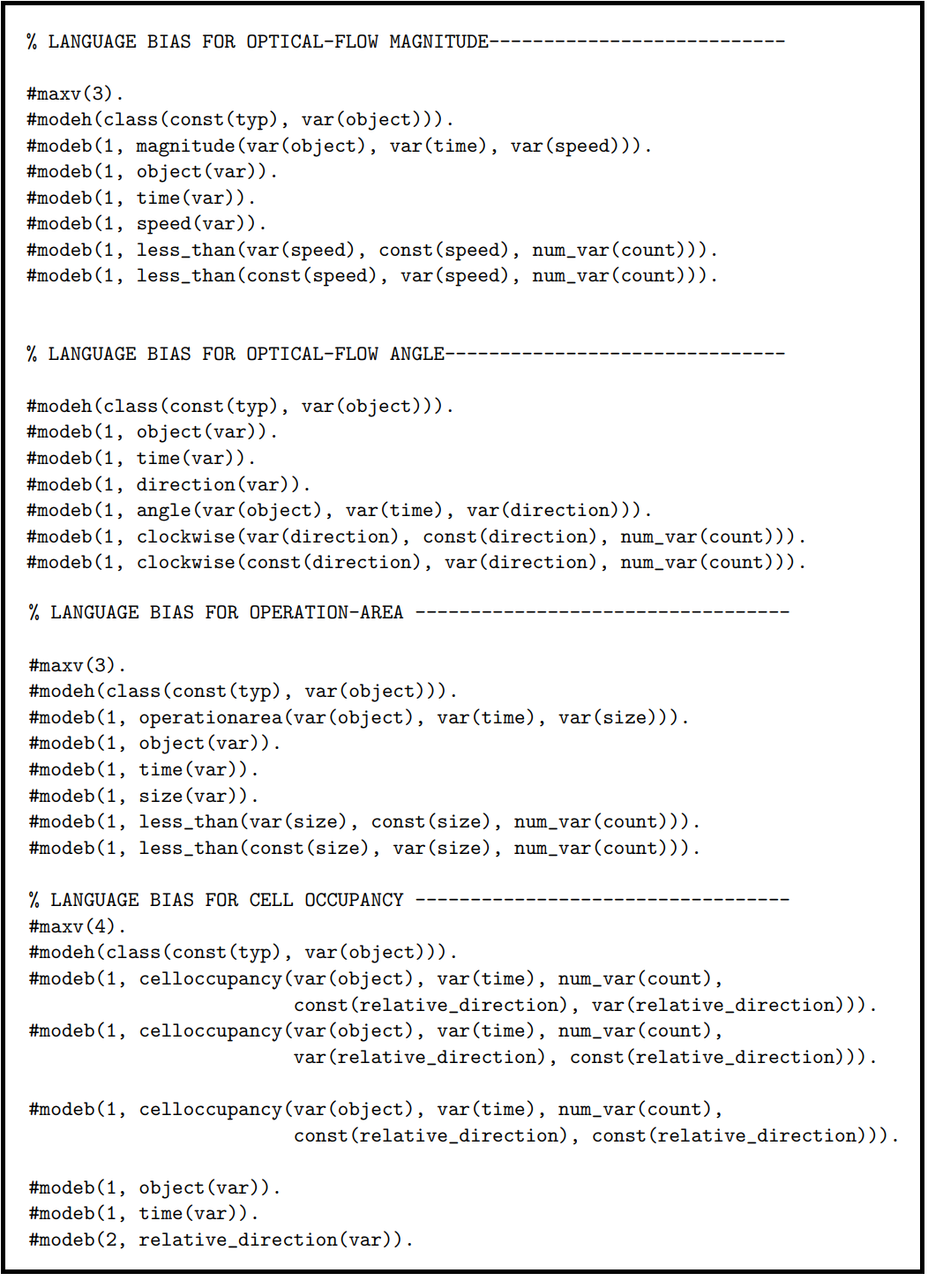}
    \caption{Language Biases used for rule-learning over optical-flow magnitude, optical-flow angle, operation-area and cell-occupancy.}
    \label{extend-lang-bias}
\end{figure}

\begin{figure}[htp]
    \centering
    \includegraphics[width=\linewidth]{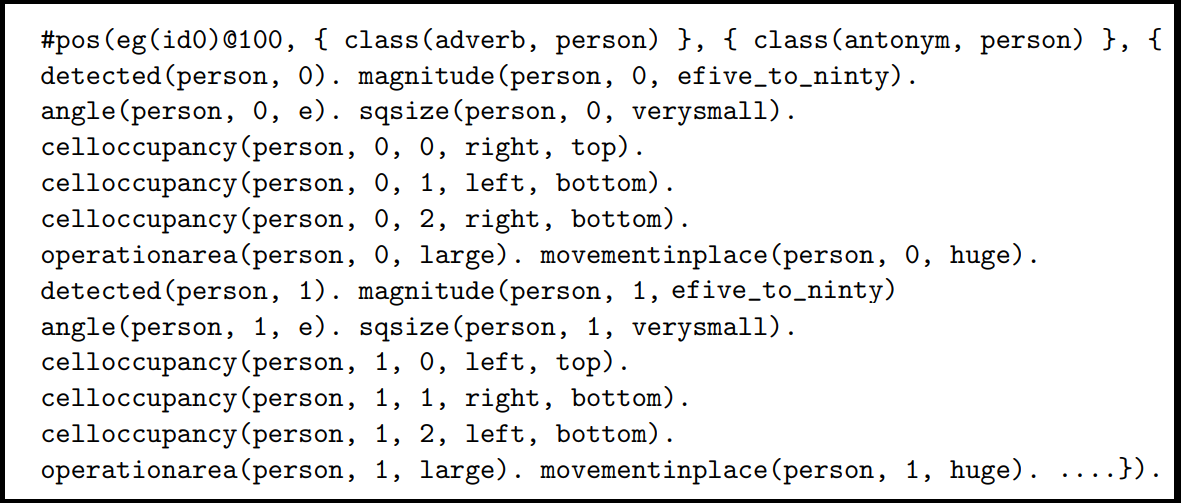}
    \caption{A truncated example of a detected object-behaviour formatted for FastLAS rule-learning.}
    \label{concat}
\end{figure}

As mentioned in Section \ref{sym-reasoning}, we consider employing the FastLAS inductive learning method \cite{law2020fastlas} to automatedly learn some governing rules over extracted object-behaviour facts - so as to explain the overall adverb-type categorization of each video clip. \\\\
In using FastLAS to learn such rules, we are primarily constrained by the number and type of variables that each rule can use. The more number of variables that a rule allows, the more the number of possible groundings that it can take, and the longer that it takes for rule learning to complete. So while each extracted instance of object behaviour possess multiple facts of the same predicates across different time-steps (such as magnitude at time step 1, magnitude at time step 2, etc.), owing to the large number of possible variable groundings across time-steps, automatedly learning rules that reason over more than one time-step is especially challenging for this task.
\\\\
In this work, rather than attempting to overcome these challenges and learn a few very complex rules over multiple time-steps and multiple predicate-properties, we instead consider learning a large number of simpler rules, that compositionally might inform overall adverb-type.
\\\\
To limit the number of free variables that FastLAS has to deal with, we focus on learning range-rules that define upper or lower bounds of an object's predicate-properties at single time-steps. To illustrate this idea, Figure \ref{toy-eg} shows a toy example that we feed into FastLAS. The example specifies the behavior four objects: a car, a plane, a person, and a cat. Each of these behaviors has a corresponding optical flow magnitude for an arbitrary time-step and each object-behaviour (that is a positive example for our rule-learning problem) is also associated with a particular class type - either `strange' or `not-strange'. The first class-type mentioned in a $\#pos$ header in the figure is the one that we wish to associate with the object-behaviour. The second class type mentioned in the header is what the object-behaviour is \underline{not}.\\\\
As shown in this toy example, in order to reduce the number of variable-values that FastLAS deals with, we also use discretized versions of optical-flow magnitude such as `five-to-ten', `ten-to-fifteen', etc.
\\\\
The language-bias shown in the figure specifies that the head atom of any learnt rule must be a class type (`strange' or `not-strange'), and must generalize over variable objects-types. The language-bias also specifies that body atoms (if used) must capture some range property over magnitude. As FastLAS can learn to use one or none of each of the specified `less-than' body atoms, a learnt-rule might enforce an upper bound on magnitude value, a lower bound on magnitude value, or neither. Additionally, as the `number-of-steps' field in the less-than predicate is specified as a  FastLAS numeric-variable (num-var), FastLAS is allowed to learn  numeric-constraints that further explain range-rules.
\\\\
For this particular toy example, we can explain all of the provided object-behaviours by deeming magnitudes between 5 and 20 to correspond to the class `strange' and other magnitude values to correspond to the class `not-strange'. FastLAS does infact discover such corresponding rules, as shown in Figure \ref{toy-eg-rulesA}. Figure \ref{rangesA} shows a depiction of these learnt ranges for better clarity.
\\
We can similarly employ these range-style language-biases for other predicate properties such as optical-flow angle, and operation area. Figure \ref{extend-lang-bias} shows how we might do so. As also shown in Figure \ref{extend-lang-bias}, for cell-occupancy we consider using a slightly different language bias - we allow for rule-body conditions that consist of: (A)A variable relative direction along the vertical (top/bottom) and a constant horizontal direction (left/right), (B)A variable direction along the horizontal (left/right) and a constant vertical direction (top/bottom) or (C)Both horizontal and vertical relative directions  specified as constants. We also use a numeric-variable (num-var) value for the level of cell-hierarchy used by a rule, so that we can learn rules that apply to different levels.
\\\\
An important facet of this type of observational-predicate rule learning is that it specifically requires numeric rule learning (either to learn constraints over the number-of-steps range property or the level-of-hierarchy as described), and we have chosen to employ FastLAS as it is the only framework that allows this type of automated numeric-rule learning over ASP programs.  
\\\\
Generalizing our toy example for the more complex problem-setting of recognizing adverb-types from object-behaviours is straightforward. We use recorded object behaviours from our video-to-ASP pipeline as positive examples in the rule-learning setup, wherein the class associated with each object-behaviour is the ground-truth adverb-type of the overall video clip that an object hails from. Naturally, the class \underline{not} to be associated with each object behaviour is the antonym of its adverb-type. Figure \ref{concat} shows a truncated example of a detected object's behaviour formatted for FastLAS rule-learning. To simplify our explorations and to reduce noise in the training data, we ignore `unknown' type object-behaviours detected by our pipeline.\\
\\
However, problems arise in using this learning methodology as-is over sets of object-behaviours for given (adverb, antonym) pairs. Firstly if the set of object-behaviours is not-balanced to have an equal number of objects for both adverb and antonym, then we might get best coverage by just predicting a single class rule without any body conditions. This problem of unbalanced data is easily solved by repeating object-behavior examples in the training data to balance out the adverb/antonym classes. The next problem is more important - since the adverb-recognition setting is quite noisy (with many of our detected object-behaviours not necessarily impacting a video's overall adverb-type), for large sets of object-behaviour, even after balancing, we find that rules learnt from our simple language biases (Figure \ref{extend-lang-bias}) are not able to cover more than $50\%$ of behaviours. Then, directly predicting one or another adverb/antonym head with no body conditions again becomes the best strategy for maximum coverage.
\\\\
As mentioned earlier, increasing the complexity of possible rules to abate this problem comes with its own set of challenges (namely rule-learning can slow down to the point of becoming computationally impractical). So, rather than increasing the complexity of our language biases, we consider smaller 20 sample batches of balanced subsets of the training data (10 samples for an adverb and 10 samples for it's antonym) - which poses a less noisy and less complex problem setting when each batch is viewed separately. We run FastLAS separately over each balanced-batch of object behaviours to get batch-wise plausible rules for each of our language-biases. When the rules returned by FastLAS for a batch and language bias are non-trivial (possess body-conditions), we then record them as indicators of adverb-type. 
\\\\
After all such indicator-rules have been extracted from a stream of balanced-batches of the full training set for each (adverb, antonym) pair using our language-biases (Figure \ref{extend-lang-bias}), we then consider composing their results together using Support Vector Machines (SVMs). Specifically, given an input object-behaviour and an adverb vs antonym task, we assign a 1 to each corresponding indicator-rule if the rule fires for the given object's behaviour, and assign a 0 otherwise - so that from our collection of indicator-rules of the adverb/antonym task, we obtain for the object-behaviour a feature of 1s and 0s (such as [1,1,1,0,1,1,...]). The entire balanced training-set is converted in this manner for each adverb vs antonym task. A separate binary-SVM with rbf kernel is trained over these extracted features (along with action-word-embeddings) to classify between every adverb and its antonym. 
\begin{figure*}[htp]
    \centering
    \begin{subfigure}[b]{0.45\textwidth}
    \centering
    \includegraphics[width=\linewidth]{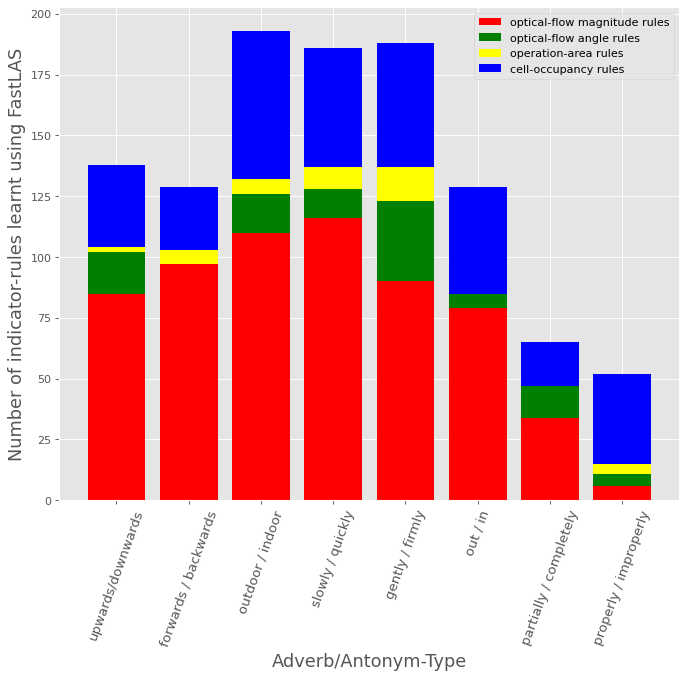}
    \caption{MSR-VTT-ASP}
    \end{subfigure}
     \begin{subfigure}[b]{0.47\textwidth}
    \centering
    \includegraphics[width=\linewidth]{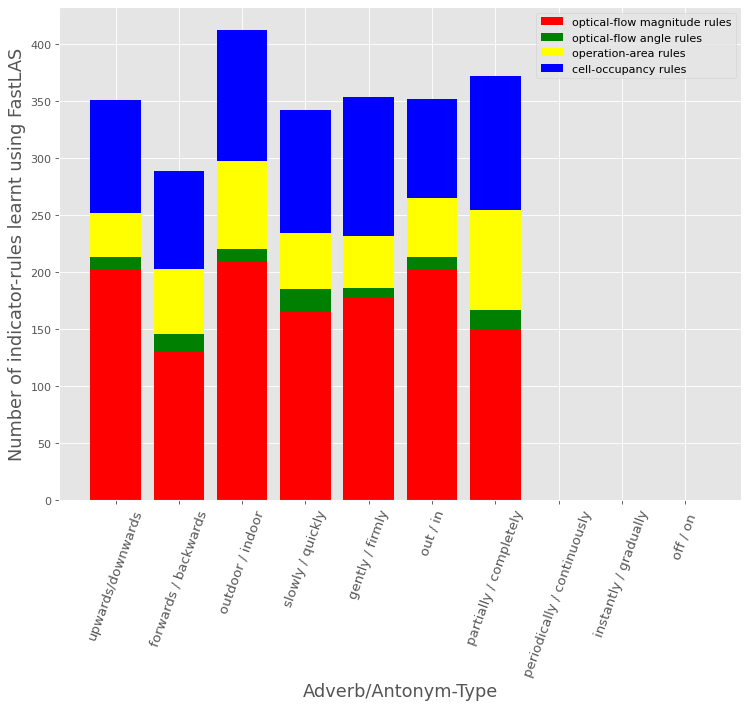}
    \caption{ActivityNet-ASP}
    \end{subfigure}
    \caption{Number of indicator-rules learnt by FastLAS across all training batches for each adverb/antonym task using our single-step magnitude, angle, operation-area and cell-occupancy language-biases (duplicates not removed).}
    \label{rules-bar-chart}
\end{figure*}\\
At inference time, a raw video clip is converted to an ASP program of object behaviors, and all the indicator-rules are checked to obtain vectors of zeros and ones for each object. All the SVMs make their adverb/antonym predictions (over features from their corresponding indicator-rules) for each detected object, and predictions from multiple objects detected within a single-clip are aggregated by a simple voting mechanism in each adverb vs antonym category. 
\subsection{Indicator Rules Learnt by FastLAS}
\begin{figure}[htp]
    \centering
    \includegraphics[width=\linewidth]{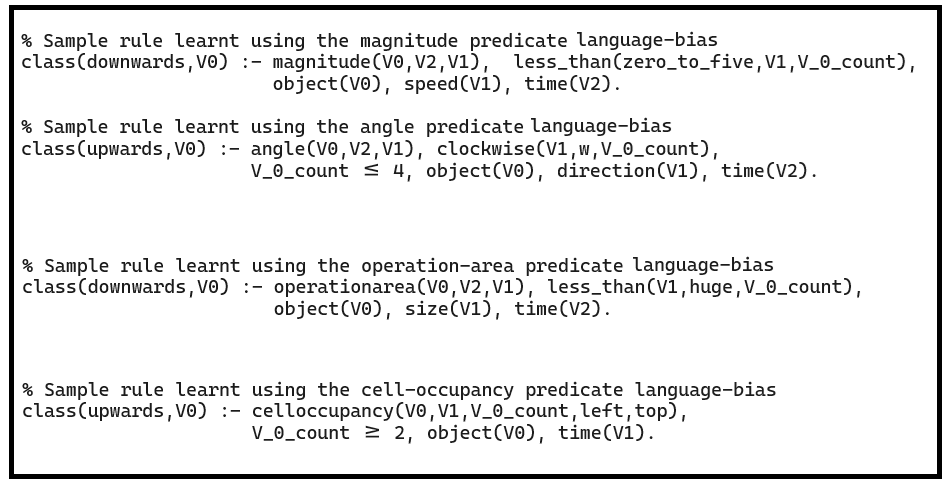}
    \caption{Some sample rules learnt by FastLAS using the magnitude, angle, operation-area and cell-occupancy language-biases specified in Figure \ref{extend-lang-bias} for some batches of examples in the upwards/downwards category.}
    \label{sample-rules-learnt}
\end{figure}
For clarity, Figure \ref{sample-rules-learnt} shows some sample rules learnt by FastLAS for each of the predicate-property types within the upward/downwards category of MSR-VTT-ASP. Note again that these rules are not necessarily universally true for the training-data, but were found by FastLAS to hold for some given batches of 10 `upward' object behaviour facts and 10 `downward' behavior facts and that we use them as indicators for a more meaningful downstream classification.\\\\
Figure \ref{rules-bar-chart} shows a count of the number and type of rules learnt across all balanced-batches of the training-data using FastLAS in each adverb/antonym category for each dataset. Note that these counts may include duplicate rules learnt, and the indicator-feature vectors for downstream classification are constructed including duplicates since the number of duplicate-rules learnt provides a rough measure of rule-strength. That is, if the same particular rule is learnt more frequently for more than one balanced batch from the training data, then it is likely a stronger indicator for that adverb-vs-antonym categorization, and we pass that information on to the downstream SVM classification by including duplicates in the feature-construction.\\\\
We highlight that for the categories of \textit{periodically / continuously, instantly /gradually and off / on}, FastLAS does not return any plausible rules. It is likely that these categories are too complex or too multi-time-step dependent for any single-time-step rules to prove sufficient explanations - even for small-batches. \\\\
Further, while the single-time-step baseline did not itself achieve state-of-the-art performance, the number of rules learnt within each type within each category does offer some insights into each adverb-vs-antonym recognition task. For example: The number of optical-flow-angle rules learnt for forward-backwards classification is zero for MSR-VTT-ASP - suggesting that angle-of-motion information does not often help determine if an object is moving forwards or backwards in that dataset; on the other hand, we have a relatively large number of operation-area based rules for outdoor-indoor categorization in ActivityNet-ASP - suggesting that perhaps objects move around more or less based on whether they are situated outdoors or indoors in ActivtityNet. Such observations could offer insights for better video-property extractions - so that one may tailor reasoning better for each adverb-vs-antonym task.   
\subsection{Compute Requirements for Experiments}
\label{fastlas-appendix}
All experiments presented in this work can be reproduced using a single P5000 GPU device. Using this resource, ASP-Program facts were extracted for the full MSR-VTT-ASP and ActivityNet-ASP datasets sequentially over 2 days, while transformer-based finetuning completes in under 1 hr for a given dataset for DistilBERT, ALBERT and BERT architecures. Learning rules from balanced-batches of object-behaviours using FastLAS and a single CPU requires roughly 20 hours to complete for each dataset. All SVM training and inference completes in under 5 minutes.


\end{document}